%% file: main.tex
\documentclass{article}
\usepackage{ijcai26}

\usepackage{times}
\usepackage{soul}
\usepackage{url}
\usepackage[hidelinks]{hyperref}
\usepackage[utf8]{inputenc}
\usepackage[small]{caption}
\usepackage{graphicx}
\usepackage{amsmath}
\usepackage{amsthm}
\usepackage{booktabs}
\usepackage{algorithm}
\usepackage{algorithmic}
\usepackage[switch]{lineno}
\usepackage{xcolor}
\usepackage{multirow}
\title{Focus-LIME: Surgical Interpretation of Long-Context Large Language Models \\via Proxy-Based Neighborhood Selection}

\author{
Junhao Liu$^1$
\and
Haonan Yu$^2$\and
Zhenyu Yan$^{2,3}$\And
Xin Zhang$^4$\thanks{Correspondence Author}\\
\affiliations
Key Lab of High Confidence Software Technologies (Peking University),\\Ministry of Education\\School of Computer Science, Peking University, Beijing 100871, China \\
\emails
\{liujunhao\textsuperscript{1}, xin\textsuperscript{4}\}@pku.edu.cn, \{a616156\textsuperscript{2}, zhenyuyan\textsuperscript{3}\}@stu.pku.edu.cn}
\input{macro.tex}
\input{math_commands.tex}

\begin{document}

\maketitle
\input{figures/example.tex}
\begin{abstract}
    As Large Language Models (LLMs) scale to handle massive context windows, achieving \textit{surgical} feature-level interpretation is essential for high-stakes tasks like legal auditing and code debugging.
    However, existing local model-agnostic explanation methods face a critical dilemma in these scenarios: feature-based methods suffer from attribution dilution due to high feature dimensionality, thus failing to provide faithful explanations.
    In this paper, we propose \textbf{Focus-LIME}, a coarse-to-fine framework designed to restore the tractability of surgical interpretation.
    Focus-LIME utilizes a proxy model to curate the perturbation neighborhood, allowing the target model to perform fine-grained attribution exclusively within the optimized context.
    Empirical evaluations on long-context benchmarks demonstrate that our method makes surgical explanations practicable and provides faithful explanations to users.
    % Furthermore, we demonstrate its practical utility in \textit{Automated Prompt Debugging}, where surgical explanations enable effective correction of model hallucinations.
\end{abstract}

\input{chapters/1.intro.tex}
\input{chapters/2.related.tex}
\input{chapters/3.approach.tex}

\input{chapters/4.experiments.tex}

\input{chapters/6.conclusion.tex}

\bibliographystyle{named}
\bibliography{vicinity}

\appendix
\section{Ethical Statement}
There are no ethical issues.

\section{LLM Usage Statement}
We did not directly use LLMs to generate any content in this paper. We use LLMs to polish the writing of this paper, and all content is reviewed and revised by the authors.
\end{document}

%% file: macro.tex
\def\AOPC{\mathrm{AOPC}}

%% file: math_commands.tex
\usepackage{amsmath,amsfonts,bm}

\def\eqref#1{equation~\ref{#1}}
\def\1{\bm{1}}

\def\va{{\bm{a}}}

\def\vx{{\bm{x}}}

\def\vz{{\bm{z}}}

\DeclareMathAlphabet{\mathsfit}{\encodingdefault}{\sfdefault}{m}{sl}
\SetMathAlphabet{\mathsfit}{bold}{\encodingdefault}{\sfdefault}{bx}{n}

\def\gV{{\mathcal{V}}}

\def\sC{{\mathbb{C}}}

\def\sF{{\mathbb{F}}}

\def\sL{{\mathbb{L}}}

\def\sN{{\mathbb{N}}}

\def\sR{{\mathbb{R}}}
\def\sS{{\mathbb{S}}}
\def\sT{{\mathbb{T}}}
\def\sU{{\mathbb{U}}}

\def\sX{{\mathbb{X}}}

\newcommand{\E}{\mathbb{E}}

\newcommand{\R}{\mathbb{R}}

\DeclareMathOperator*{\argmin}{arg\,min}

%% file: figures/example.tex
\begin{figure*}[t]
    \centering
    % 调整宽度: width=1.0\linewidth 表示占满通栏宽度
    % 如果图片留白较多，可以改为 width=0.9\linewidth
    \includegraphics[width=\linewidth]{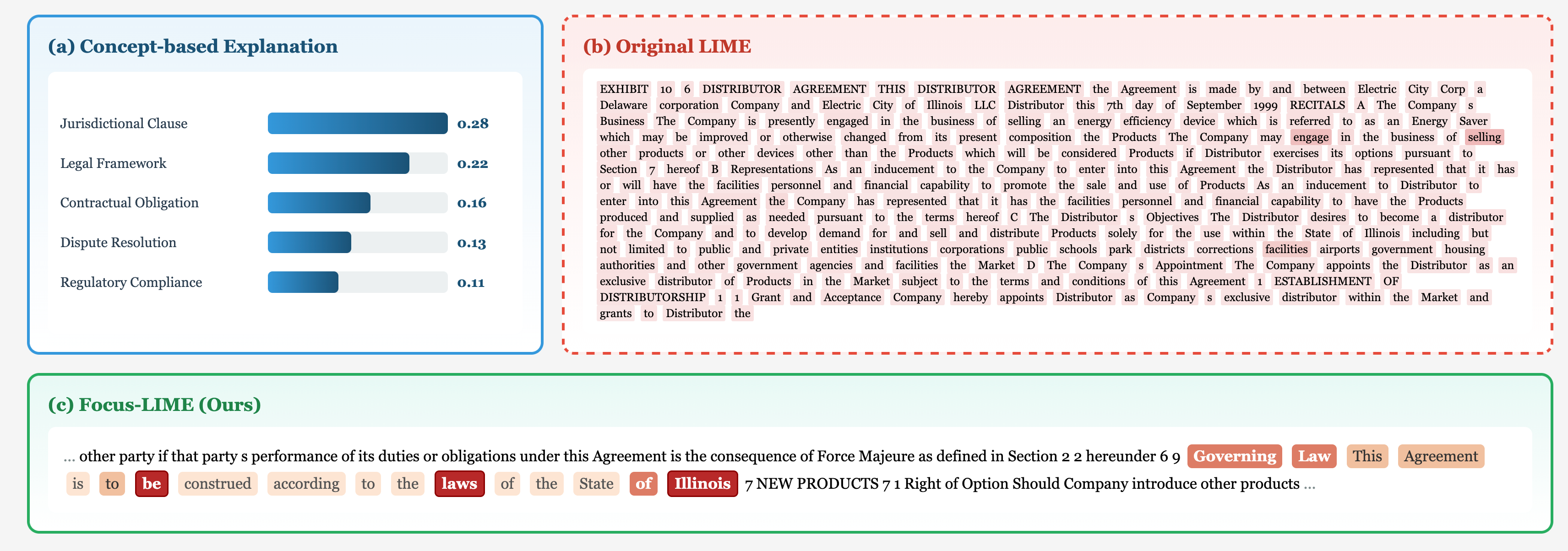}
    
    \caption{
        Comparison of explanations generated by different methods when querying LLMs for a "Governing Law" clause in a lengthy contract.
        \textbf{(A) Concept-based Explanation:} Attributes importance to high-level concepts.
        \textbf{(B) Original LIME:} Suffers from \textit{attribution dilution}, where significance scores are distributed noisily across all tokens.
        \textbf{(C) Ours (Neighborhood Reduced):} Generates a sparse and precise explanation.
    }
    \label{fig:explanation_comparison}
\end{figure*}

%% file: chapters/1.intro.tex
\section{Introduction}

As Large Language Models (LLMs) scale to handle massive context windows~\cite{ding2024longrope}, they are entrusted with increasingly complex tasks, from auditing legal contracts to debugging extensive codebases.
% especially in some professional domains, such as legal~\cite{valvoda2024towards} and medical~\cite{guo2024evaluating}.
Given the opaque nature of these models, explaining their behavior has emerged as a critical area of research~\cite{valvoda2024towards,guo2024evaluating}.
In particular, practitioners require explainability at varying levels of granularity~\cite{elisha2025rethinkingsaliencymapscognitive}.
Consequently, there is a persistent demand for faithful, fine-grained explanations that can pinpoint the exact locus of model decisions, such as locating a specific importance clause or word within a 50-page document. 
In this paper, we propose to generate surgical local explanations for LLMs by curating the neighborhood of explanations with proxy models.

To illustrate the dilemma of explaining long-context LLMs, consider the motivating example from the Contract Understanding Atticus Dataset (CUAD)~\cite{hendrycks2021cuad} shown in Figure~\ref{fig:explanation_comparison}.
Here, a user queries a 6,000-word contract to identify clauses designating the specific state or country laws applicable to the agreement.
On one hand, \textbf{concept-based explanation methods}~\cite{ACE,ConLUX,TBM} scale effectively to long contexts by operating at a high level of abstraction.
In this scenario, while they might correctly identify the model's reliance on ``Jurisdiction Clause" concepts, they inherently lack the granularity to localize specific tokens, leaving the user unable to verify the exact \textit{locus} of the decision.
\textbf{Conversely, feature-based methods} like LIME~\cite{LIME} theoretically offer this surgical precision but collapse under the \textit{curse of dimensionality}.
When applied to this extensive context, standard LIME suffers from severe \textit{attribution dilution}: it dissipates the sampling budget across thousands of irrelevant tokens, drowning the critical signal in background noise. To generate faithful explanations, LIME would require an impractically large number of model queries, also resulting in prohibitive computational costs.
Our goal is to bridge this gap by strictly curating the neighborhood, enabling the explainer to filter out semantic noise and surgically focus on the decisive evidence.
Consequently, our method can precisely identify that specific words like ``Governing", ``Law", and ``Illinois" are the key drivers of the model's decision.

To address the intractability of fine-grained attribution in long-context scenarios, we propose \textbf{Focus-LIME}, a framework designed to restore the surgical feature attribution capabilities of LIME for LLMs.
Our approach fundamentally reimagines the role of efficiency-oriented techniques.
While recent research~\cite{zhao2025leveraging,jiang2023llmlinguacompressingpromptsaccelerated,liu2025towards} has explored using Proxy Models to generate explanations entirely to reduce inference costs, this strategy inherently risks the unfaithfulness of the explanations.
Acknowledging this limitation, we draw inspiration from proxy-based efficiency but shift the paradigm from ``replacement" to ``guidance."
We employ the proxy model strictly as a ``preliminary scout" for Neighborhood Selection, tasked with filtering out irrelevant background features.
Crucially, we then perform the final perturbation analysis using the Target Model, but exclusively within this optimized, low-dimensional neighborhood.
This ``Scout then Scrutinize" strategy aims to preserve fidelity to the target model's decision boundary, while obtaining computational scalability from proxy-based methods.

We extensively evaluate Focus-LIME on long-context benchmarks, including the CUAD and Qasper~\cite{qasper} datasets.
Empirical results demonstrate that our method significantly outperforms both original LIME and pure-proxy baselines in terms of explanation fidelity.
Comparing to curating neighborhood using the target models, Focus-LIME achieves comparable fidelity while requiring notably fewer model queries, showcasing its efficiency.
%
% Beyond metrics, we demonstrate the practical utility of our surgical explanations in \textbf{Automated Prompt Debugging}: by identifying the precise locus of errors, our method enables automatic constraint re-weighting that successfully corrects model hallucinations in complex contracts.

In summary, our contributions are as follows:
\begin{itemize}
    \item We propose Focus-LIME, a model-agnostic framework that enables \textit{surgical} feature-attribution interpretation for long-context LLMs by decoupling neighborhood selection (via proxy) from attribution (via target).
    \item We instantiate this framework by designing a Coarse-to-Fine Neighborhood Contraction algorithm that effectively leverages proxy models to filter semantic noise without sacrificing target fidelity.
    \item We evaluate the effectiveness of our approach on long-context benchmarks, showing that it makes surgical explanations practicable and provides faithful explanations to users.
\end{itemize}

%% file: chapters/2.related.tex
\section{Background and Related Work}

In this section, we introduce the necessary background and survey prior studies relevant to our work.
\subsection{Large Language Models}

A large language model can be regarded as a probabilistic function \( f \) that, given a sequence of tokens \(\vx = [x_1, x_2, \ldots, x_t]\), produces a probability distribution over the set of possible next tokens, written as \( f(\vx) \).
Formally, the mapping is \( f: \gV^* \rightarrow \sR^{|\gV|} \), where \( \gV \) denotes the vocabulary, and the output corresponds to a distribution over all vocabulary elements.
Since most LLMs operate with an upper bound on input length \( n \), we typically assume the sequence \( \vx \in \sX^n \).
Nowadays, LLMs are used in a wide range of applications, including medical~\cite{guo2024evaluating} and legal~\cite{valvoda2024towards}, which often involve processing long document inputs and require reliable explanations to ensure their trustworthiness.

\subsection{Local Model-Agnostic Explanation Techniques}

A model-agnostic local explanation method \( t \) takes as input a predictive model \( f \) together with a specific instance \( \vx \), and produces an explanation \( g_{f,\vx} \) that characterizes how the model behaves around \( \vx \).

In this work, our attention is mainly on attribution-style approaches, which are the most widely used explanation form. These methods assign numerical relevance scores to the input features of \( \vx \in \sX \), aiming to capture how much each feature contributes to the output \( f(\vx) \).
The explanation \( g_{f,\vx} \) can thus be represented as a vector of attributions \( \va = [a_1, a_2, \ldots, a_n] \), where each coefficient \( a_i \) quantifies the effect of feature \( x_i \) on the model’s decision.
Formally, we can describe an attribution-based explanation method \( t \) as a function
\[
t: \sF \times \sX^n \rightarrow \R^n,
\]
where \( \sF \) denotes the set of predictive models, \( \sX^n \) is the input domain, and \( n \) is the dimensionality of the instance \( \vx \).

Local explanations describe the target model's behavior in a local neighborhood around a specific input~\cite{survey_XAI}. Specifically, the neighborhood is usually defined as a set of perturbed samples generated by perturbing the input~\cite{LIME,SHAP,survey_XAI}.
Specifically, perturbations are typically represented in a binary masking space, where each perturbed instance is encoded by a vector \(\vz \in \{0,1\}^n\).
Each entry \(z_i\) indicates whether the corresponding feature \(x_i\) of the original input \(\vx\) is preserved \((z_i=1)\) or removed/replaced \((z_i=0)\).
Formally, there exists a mapping function
\[
\phi: \{0,1\}^n \times \sX^n \rightarrow \sX^n,
\]
that reconstructs a perturbed sample in the original input space.
The neighborhood \(\sN\) of the input \(\vx\) is then defined as
\[
\sN = \{\phi(\vz, \vx) | \vz \in \{0,1\}^n\}.
\]

When processing long sentence inputs, the number of features \( n \) can be very large, leading to a large and high-dimensional neighborhood.
Many methods have been proposed to handle this challenge. Some methods~\cite{speedlime,achtibat2024attnlrpattentionawarelayerwiserelevance} use attention to narrow the neighborhood, while they require access to the internal model structure, which is not applicable to closed-source LLMs.
Some others methods~\cite{Shankaranarayana_Runje_2019,tan2023glime} try to improve the sampling strategy, while they still provide explanations in a large neighborhood, which can lead to unfaithful explanations.

\subsection{Desiderata of Explanations}

In this paper, we focus on the fidelity and cost of explanations.~\cite{yeh2019fidelity,burger2023your}.

On one hand, explanations should faithfully reflect the model's decision-making process. High fidelity indicates that the explanation accurately captures how the model arrives at its predictions.
This ensures that users can trust the explanation and apply it to various practical scenarios, such as model debugging and bias detection.

On the other hand, when explaining large language models, the cost of generating explanations becomes a critical consideration, as querying these models can be computationally expensive.
Therefore, it is essential to develop explanation methods that can deliver high-fidelity explanations while adhering to practical computational budgets.
In this work, we aim to improve the fidelity of explanations without increasing the overall computational cost.

%% file: chapters/3.approach.tex
\section{Methodology}
\input{figures/pipeline.tex}
In this section, we introduce the Focus-LIME framework for generating surgical explanations for long-context LLMs via proxy-based neighborhood selection. 

\subsection{Problem Formulation}
Following the standard setting of local model-agnostic attribution-based explanation methods~\cite{LIME,SHAP}, the goal is to estimate the optimal attribution \(\va*\) vector that minimizes the local approximation error \(L\) over the neighborhood:
\begin{equation}
\va^* = \argmin_{\va} \E_{\vz \sim \pi_{\vx}} \left[ L(f_T(\phi(\vz, \vx)), g_{\va}(\vz)) \right] + \Omega(\va),
\label{eq:standard_lime}
\end{equation}
where \(\pi_{\vx}\) is the sampling distribution and \(\Omega(\va)\) is a complexity regularizer .

\paragraph{The Cost-Fidelity Dilemma.}
The fidelity of the local explanation relies heavily on the \textit{sample density} within the neighborhood.
To learn a faithful linear surrogate \( g_{\va} \), the number of perturbed samples \( K \) must scale with the feature dimension \( n \) to ensure the regression stabilizes.

However, as querying large language models incurs significant computational costs, for end-users, there is often a budget \( B \) that limits the total number of tokens that can be processed.
Under a computational budget \( B \), the maximum allowable samples \( K_{max} = \lfloor B / C(f_T) \rfloor \) is fixed.
In long-context scenarios, as \( n \) grows, the sample density \( \rho = K_{max} / n \) becomes vanishingly small thus the estimation of \( \va* \) becomes increasingly unreliable.
The limited samples fail to cover the high-dimensional decision boundary, causing the variance of the estimator \( \va^* \) to explode.
We term this phenomenon \textit{attribution dilution}, where the resulting explanation becomes indistinguishable from noise.

\paragraph{Proposed Formulation.}
To resolve this, we introduce a computationally efficient proxy model \(f_P\) (where \(C(f_P) \ll C(f_T)\)) and decompose the problem into two stages.
Instead of optimizing over the entire neighborhood \(\sN\), we seek to identify an \textit{active neighborhood subspace} \(\sS \subseteq \sN\) that contains the most information.
Our objective is reformulated as:
\begin{equation}
\va^*_{focused} = \argmin_{\va} \E_{\vz \sim \pi_{\vx}'} \left[ \sL(f_T(\phi(\vz, \vx)), g_{\va}(\vz)) \right]+ \Omega(\va),
\label{eq:focus_lime}
\end{equation}
where the subspace \(\sS\) is determined by the proxy \(f_P\) to filter out irrelevant dimensions (semantic noise), allowing the target model \(f_T\) to focus its limited sampling budget on the critical decision boundary. In this case, the sample density \(\rho' = K_{max} / |\sS|\) can be maintained at a reasonable level, enabling faithful estimation of \(\va^*_{focused}\).

\subsection{The Focus-LIME Framework}
\label{sec:framework}

To address the dilemma of vanishing sample density, we propose \textbf{Focus-LIME}, a framework that implements a coarse-to-fine neighborhood curation strategy.
The overall pipeline is illustrated in Figure~\ref{fig:pipeline}. 
Our core insight is to \textbf{decompose} the explanation process into two distinct phases, using a proxy model to first \textit{scout} the relevant neighborhood, which does not require additional target model queries; followed by using the target model to \textit{scrutinize} the fine-grained attributions within this focused region.

\paragraph{Phase I: Proxy-Guided Scouting.}
This phase aims to identify an \textit{Active Neighborhood Subspace} \( \sS \subset \sN \) that contains the most relevant features, thereby filtering out background noise.
We formulate Focus-LIME as a flexible framework where the \textit{scouting policy} is a modular component. Depending on the nature of the proxy model and the task, various selection strategies can be employed (e.g., gradient-based saliency~\cite{IG}, attention~\cite{speedlime,achtibat2024attnlrpattentionawarelayerwiserelevance}, and so on).
In our experiments, we instantiate this as an \textbf{iterative refinement process} in a fully model-agnostic manner, structured as follows:

This process proceeds as follows:

\begin{enumerate}
    \item \textbf{Initialization:} We initialize the candidate set \( \sC^{(0)} \) as the entire input document \( X \). Let \( t = 1 \) be the current iteration step.

    \item \textbf{Hierarchical Decomposition:} At step \( t \), we decompose the surviving candidates from the previous step \( \sC^{(t-1)} \) into finer-grained units \( \sU^{(t)} = \{u_1, u_2, \dots, u_m\} \).
    (e.g., In the first iteration, we decompose the document into paragraphs; in the second, we decompose the selected paragraphs into sentences).

    \item \textbf{Proxy Valuation:} We employ LIME to compute a attribution score for each unit in \( \sU^{(t)} \) using the proxy model \( f_P \). We follow ~\cite{liu2025towards} to choose the proxy model.

    \item \textbf{Top-\(k\) Filtering:} We select the top-\( k_t \) units with the highest relevance scores to form the refined set \( \sC^{(t)} \).

    \item \textbf{Termination Check:} We check if the total token length of \( \sC^{(t)} \) satisfies the density constraint or a user-defined stop condition (e.g., maximum number of iterations).
    If the constraint is met, the iteration terminates, and \( \sC^{(t)} \) is converted into the final binary \textit{Focus Mask} \( \mathbf{m}_{focus} \). Otherwise, we increment \( t \) and repeat from Step 2. 
\end{enumerate}

Through this iterative pruning, Focus-LIME effectively ``zooms in" on the decision boundary, discarding massive amounts of irrelevant context.
\paragraph{Phase II: Target-Based Scrutinizing.}
Once the iterative scouting terminates, we obtain a converged \textit{Focus Mask} \( \mathbf{m}_{focus} \in \{0, 1\}^n \), which delineates the important region from the irrelevant context.
In this phase, we engage the target model \( f_T \) to perform fine-grained attribution.
Unlike standard LIME, which suffers from dilution by sampling uniformly across the entire high-dimensional space, Focus-LIME restricts the perturbation logic as follows:

\begin{enumerate}
    \item \textbf{Constrained Perturbation:} We generate a set of \( K \) perturbed samples \( \mathcal{Z}' = \{z'_1, \dots, z'_K\} \) using a \textit{Mask-Conditioned Generator}.
    For each sample \( z' \in \{0, 1\}^n \), the value of the \( i \)-th token \( z'_i \) is determined by:
    \begin{equation}
        z'_i = 
        \begin{cases} 
        1 (Static), & \text{if } \mathbf{m}_{focus}[i] = 0 \\
        \sim \text{Bernoulli}(0.5), & \text{if } \mathbf{m}_{focus}[i] = 1 
        \end{cases}
    \end{equation}
    This strategy ensures that the irrelevant context remains static, while the perturbation is concentrated exclusively within the active region identified using the proxy model.

    \item \textbf{Target Valuation:} We query the expensive target model \( f_T \) with these constrained samples to obtain the prediction vector \( \mathbf{y}' \).
    Crucially, because the dimensionality of the perturbation space has been reduced from \( n \) (total tokens) to \( n_{active} \), where \( n_{active}:= \sum_{i=1}^n \mathbf{m}_{focus}[i] \), the sample density \( \rho = K / n_{active} \) is restored to a high level, ensuring the stability of the subsequent regression.

    \item \textbf{Explanation Generation:} Finally, following the standard LIME procedure, we fit a weighted linear regression model to approximate the target model's behavior within the focused neighborhood.
    \begin{equation}
        \phi^* = \operatorname*{argmin}_{\phi} \sum_{z' \in \mathcal{Z}'} \pi_{\sS}(z') \cdot \mathcal{L}(f_T(z'), \phi^\top z') + \Omega(\phi)
    \end{equation}
    where only the weights corresponding to the active region (\( \mathbf{m}_{focus}[i] = 1 \)) are optimized.
    The resulting attribution vector \( \phi^* \) provides a \textit{surgical explanation}: it identifies the precise tokens that most significantly influenced the target model's prediction  within the long context input.
\end{enumerate}

\subsection{Summary}
\label{sec:method_summary}

\input{tables/fidelity_results.tex}
In summary, we presented \textbf{Focus-LIME}, a framework designed to overcome the intractability of explaining long-context LLMs.
By recognizing that not all tokens in a massive document require equal scrutiny, we reformulated the local explanation problem as a two-stage process:
\begin{itemize}
    \item \textbf{Scout (Phase I):} Utilizing a cost-efficient proxy model to iteratively filter out semantic noise and identify a low-dimensional \textit{Active Neighborhood}.
    \item \textbf{Scrutinize (Phase II):} Leveraging the target model to perform surgical attribution exclusively within this optimized subspace.
\end{itemize}
This approach resolves the fundamental conflict between \textit{attribution dilution} and \textit{computational budget} in existing model-agnostic feature-attribution methods.
This ensures that the final explanation achieves high fidelity to the target model's behavior while keeping the inference cost manageable.

%% file: figures/pipeline.tex
\begin{figure*}[t]
    \centering
    % 调整宽度: width=1.0\linewidth 表示占满通栏宽度
    % 如果图片留白较多，可以改为 width=0.9\linewidth
    \includegraphics[width=\linewidth]{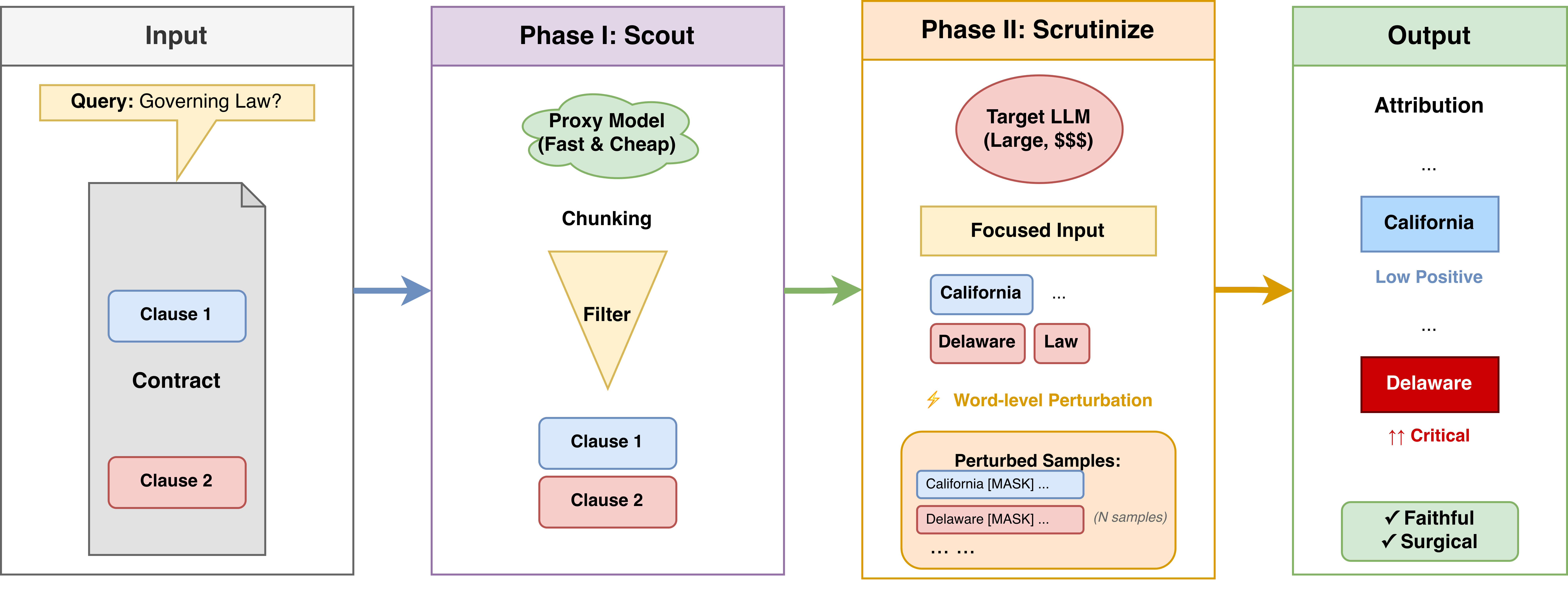}
    
    \caption{
        The pipeline of Focus-LIME, which consists of two main steps: 1) Neighborhood Curation with Proxy Models: we first use a smaller proxy model to iteratively narrow the neighborhood by fixing unimportant features, and identify the most faithful neighborhood; 2) Fine-grained Attribution with Target Model: we then generate the final explanations by using the original LLM, but only in the curated neighborhood.
    }
    \label{fig:pipeline}
\end{figure*}

%% file: tables/fidelity_results.tex
\begin{table*}[t]
\centering
\caption{Faithfulness evaluation on \textbf{CUAD} measured by AOPC. Higher is better.}
\label{tab:faithfulness_cuad}
\setlength{\tabcolsep}{6pt}
\small
\begin{tabular}{l l c c c c}
\toprule
Target Model & Method 
& $AOPC_{10}$ & $AOPC_{50}$ & $AOPC_{100}$ & $AOPC$ \\
\midrule

\multirow{5}{*}{GPT-4o}
& Original LIME               & 0 & 0 & 0 & 0 \\
& Proxy Explanation (Qwen3-30B-A3B)           & 0 & 0 & 0 & 0 \\
& Focus-LIME (w/o Proxy)     & 0.214 & 0.433 & 0.761 &  0.489 \\
& \textbf{Focus-LIME (Qwen3-30B-A3B)}    & 0.221 & 0.452  &  0.753 & 0.481 \\
& \textbf{Focus-LIME (Qwen2.5-32B)}     & 0.207 & 0.431 & 0.757 &  0.473 \\
\midrule

\multirow{5}{*}{DeepSeek V3}
& Original LIME               & 0 & 0 & 0 & 0 \\
& Proxy Explanation (Qwen3-30B-A3B)           & 0 & 0 & 0 & 0 \\
& Focus-LIME (w/o Proxy)      &0.143 & 0.561 & 0.786 & 0.535  \\
& \textbf{Focus-LIME (Qwen3-30B-A3B)}     & 0.164 & 0.546  & 0.775 & 0.524 \\
& \textbf{Focus-LIME (Qwen2.5-32B)}       & 0.157 & 0.537 & 0.773 & 0.531 \\
\bottomrule
\end{tabular}
\end{table*}

\begin{table*}[t]
\centering
\caption{Faithfulness evaluation on \textbf{Qasper} measured by AOPC. Higher is better.}
\label{tab:faithfulness_qasper}
\small
\begin{tabular}{l l c c c c}
\toprule
Target Model & Method 
& $AOPC_{10}$ & $AOPC_{50}$ & $AOPC_{100}$ & $AOPC$ \\
\midrule

\multirow{5}{*}{GPT-4o}
& Original LIME               & 0 & 0 & 0 & 0 \\
& Proxy Explanation (Qwen3-30B-A3B)           & 0 & 0 & 0 & 0 \\
& Focus-LIME (w/o Proxy)      & 0.083 & 0.357 & 0.413 &  0.291 \\
& \textbf{Focus-LIME (Qwen3-30B-A3B)}        & 0.102  & 0.356 & 0.408 & 0.287 \\
& \textbf{Focus-LIME (Qwen2.5-32B)}          & 0.081 & 0.341 & 0.391  & 0.279 \\
\midrule

\multirow{5}{*}{DeepSeek V3}
& Original LIME               & 0 & 0 & 0 & 0 \\
& Proxy Explanation (Qwen3-30B-A3B)           & 0 & 0 & 0 & 0 \\
& Focus-LIME (w/o Proxy)      & 0.064 & 0.276 & 0.384 & 0.273 \\
& \textbf{Focus-LIME (Qwen3-30B-A3B)}          & 0.074 & 0.265 & 0.375 & 0.287 \\
& \textbf{Focus-LIME (Qwen2.5-32B)}          & 0.068 & 0.258 & 0.369 & 0.279 \\
\bottomrule
\end{tabular}
\end{table*}

%% file: chapters/4.experiments.tex
\section{Empirical Study}
In this chapter, we present a comprehensive empirical evaluation of \textbf{Focus-LIME} on challenging long-context benchmarks.
Concisely, we aim to answer the following research questions:
\begin{itemize}
    \item \textbf{RQ1 (Faithfulness):} Does Focus-LIME generate more faithful explanations compared to original methods (e.g., LIME) and pure-proxy baselines in long-context scenarios?
    
    \item \textbf{RQ2 (Mechanism Verification):} Does neighborhood curation process mitigate attribution dilution thus improving explanation fidelity?
    
    \item \textbf{RQ3 (Human Alignment):} Does Focus-LIME effectively align with human reasoning by accurately localizing the ground-truth evidence spans annotated by experts?
\end{itemize}

\subsection{RQ1: Faithfulness Evaluation}
\subsubsection{Experimental Setup}
\paragraph{Datasets.}
We use two long-context document QA datasets in our experiments:
\begin{itemize}
    \item \textbf{CUAD}~\cite{hendrycks2021cuad} is a legal document QA dataset that requires models to answer questions based on long legal contracts. 
    \item \textbf{Qasper}~\cite{qasper} is a long-context QA dataset that contains scientific papers.
\end{itemize}
We use the yes/no question answering tasks in both datasets for evaluation.

\paragraph{Models.}
We evaluate our method on different combinations of target and proxy models:
\begin{itemize} 
    \item \textbf{Target Models ($f_T$):} We use GPT-4o~\cite{GPT4} and DeepSeek V3 as target models.
    \item \textbf{Proxy Models ($f_P$):} After the screening process, we use Qwen3-30B-A3B and Qwen2.5-14B models as the proxy model for all target LLMs, which can run on consumer-grade GPUs.
\end{itemize}

\paragraph{Baselines.}
We compare Focus-LIME against the following baselines: 
\begin{itemize}
    \item \textbf{Original LIME:} Applies LIME directly to the full document. For each test, we sample 1000 perturbed samples.
    \item \textbf{Proxy Explanation:} Following~\cite{liu2025towards}, uses the proxy model to generate explanations. As proxy is much cheaper than the target model, we sample 10,000 perturbed samples for each test.
    \item \textbf{Focus-LIME (w/o Proxy Model):} An ablation of Focus-LIME that skips the proxy-based neighborhood curation step, directly applying constrained perturbation on the full document.
\end{itemize}

\paragraph{Metrics.} To evaluate the faithfulness of explanations, we conduct \textbf{deletion experiments}~\cite{LIME,SHAP,EAC}, which measure the change in model output when the most important features (as identified by the explanation) are removed. We used \textit{Area Over most relevant first perturbation curve} (AOPC) as the metric. Specifically, AOPC is defined as:
\[
        \AOPC_k = \frac{1}{|\sT|} \sum_{\vx \in \sT} (p_f(y|\vx)-p_f(y|\vx^{(k)})), 
\]
where $p_f(y|\vx)$ is the probability of $f$ to output $y$ given the input $\vx$, and $\sT$ is the set of all test inputs, and $\vx^{(k)}$ is the input $\vx$ with the top $k$ most important features removed.
A higher $\AOPC_k$ indicates a better explanation.
We report \(AOPC_{10}\), \(AOPC_{50}\), and \(AOPC_{100}\) and \(AOPC:=\frac{\sum_{k=1}^{100} AOPC_k}{100}\) in our experiments.

\subsubsection{Results}
Tables~\ref{tab:faithfulness_cuad} and~\ref{tab:faithfulness_qasper} present the faithfulness evaluation results on the CUAD and Qasper datasets, respectively. 
We observe that in the \textbf{absence of neighborhood curation}, both Standard LIME and Pure Proxy baselines struggle significantly in long-context scenarios. Specifically, they yield negligible AOPC scores, failing to identify truly decisive features due to severe \textit{attribution dilution}. 
In contrast, by incorporating the neighborhood curation process, Focus-LIME restores the tractability of the explanation task, making it practical and useful by substantially enhancing fidelity.

On the other hand, Focus-LIME have comparable performance to the ablation without proxy model, which indicates that using proxy models for neighborhood curation does not sacrifice the fidelity of explanations. Additionally, Focus-LIME using different proxy models (Qwen3-30B-A3B and Qwen2.5-32B) achieve similar performance, which indicates that our method is robust to the choice of proxy models.

\input{figures/rq2_1.tex}
\input{figures/rq2_2.tex}
\subsection{RQ2: Mechanism Verification}

While RQ1 has demonstrated the superior faithfulness of Focus-LIME on long-context benchmarks, we further investigate the underlying mechanism by verifying our hypothesis that \textbf{narrowing the neighborhood can improve the fidelity of local} explanations in general scenarios, thus users can also utilize this strategy to improve the fidelity of explanations on not only long-context scenarios.

\paragraph{Experimental Setup.}
We conducted an experiment on two text classification datasets: the Large Movie Review Dataset (IMDb)~\cite{IMDB} datasets, which contains movie reviews with average length of 228 words.
We use Qwen3-235B-A22B as the target model, and use Qwen3-30B-A3B as the proxy model to optimize the neighborhood.
We optimze the neighborhood by greedily fixing the least important feature according to the explanation generated by LIME in each iteration, and evaluate the fidelity of explanations on the path of narrowing the neighborhood.
We use AOPC as the fidelity metric, and compare the fidelity of explanations on the original neighborhood and the optimal neighborhood found on the path of narrowing the neighborhood. 

\paragraph{Results.}

% \paragraph{Results}

Figure~\ref{fig:fid} shows the deletion fidelity (AOPC) of explanations on the original neighborhood and optimal neighborhood from deleting 1 to 100 most important features. 
On average, using the optimal neighborhood can relatively improve the fidelity by 11.4\%. 
When deleting 10 most important features, using the optimal neighborhood can relatively improve the fidelity by 32.2\%.

Moreover, Figure~\ref{fig:imp2} provides a more fine-grained analysis of the fidelity improvement.
Figure~\ref{fig:imp2} (a) provides a overview of the relation between the improvement of fidelity, the number of deleted features, and the percentage of examples.
Specifically, Figure~\ref{fig:imp2}(b) shows that over 20\% of the examples can achieve more than 10\% relative improvement in fidelity by using the optimal neighborhood across different deletion ratios, and when deleting less than 10 features, over 50\% of the examples can achieve more than 10\% relative improvement in fidelity.
Figure~\ref{fig:imp2}(c) further shows that using the optimal neighborhood can significantly improve the fidelity for examples, especially when deleting a small number of features.

% \subsection{Evaluation of Our Method}
% \paragraph{Setup}
% % We still focused on the text classification task and used the Large Movie Review Dataset (IMDb)~\cite{IMDB}.
% %
% % We used the Bert model~\cite{devlin2018bert} as the proxy model to optimize the neighborhood, and check the fidelity improvement on three Qwen2.5 models~\cite{qwen25} with different parameter sizes: 3B, 14B, and 72B.
% We conducted deletion results to compare the explanations on the original neighborhood and the neighborhood optimized by our method, and also used AOPC as the metric.

% \subsection{Results}
% We chose the fidelity of deleting the 20 most important features as an example. The results show that using the optimized neighborhood can improve the fidelity on 100\% of the sentences, and the fidelity is relatively improved by 129.8\% for the 3B model, 190.8\% for the 14B model, and 259.4\% for the 72B model on average.

\subsection{RQ3: Alignment with Human Evidence}
\label{sec:human_alignment}
\input{figures/case_study.tex}
RQ3 investigates the \textit{practical utility} of Focus-LIME: can it serve as a reliable tool for human users to locate critical information?
To answer this, we utilize the ground-truth annotations provided in the CUAD dataset.
A high-quality explanation for a robust model should align with these human annotations.

\paragraph{Quantitative Analysis: Evidence Localization.}
We treat the top-\(k\) words identified by the explainer as a "retrieval set" and evaluate their overlap with the human ground truth using \textbf{Recall}.
As the length of human-annotated evidence varies across examples, we choose \(k\) as a relative ratio of the evidence length, specifically \{100\%, 150\%, 200\%\}\ of the total number of words in the human-annotated evidence.
In this experiment, we follow the setting of RQ1, using GPT-4o as the target model and Qwen3-30B as the proxy model.

As shown in Table~\ref{tab:alignment}, our methods can effectively localize the ground-truth evidence spans annotated by experts. In Figure~\ref{fig:case-study}, we present a example where Focus-LIME successfully identifies the critical clauses that determine the model's decision, demonstrating its practical utility in assisting users to navigate and verify long documents.
\input{tables/human_alignment.tex}

%% file: figures/rq2_1.tex
\begin{figure}
    \centering
    \resizebox{0.4\textwidth}{!}{\includegraphics{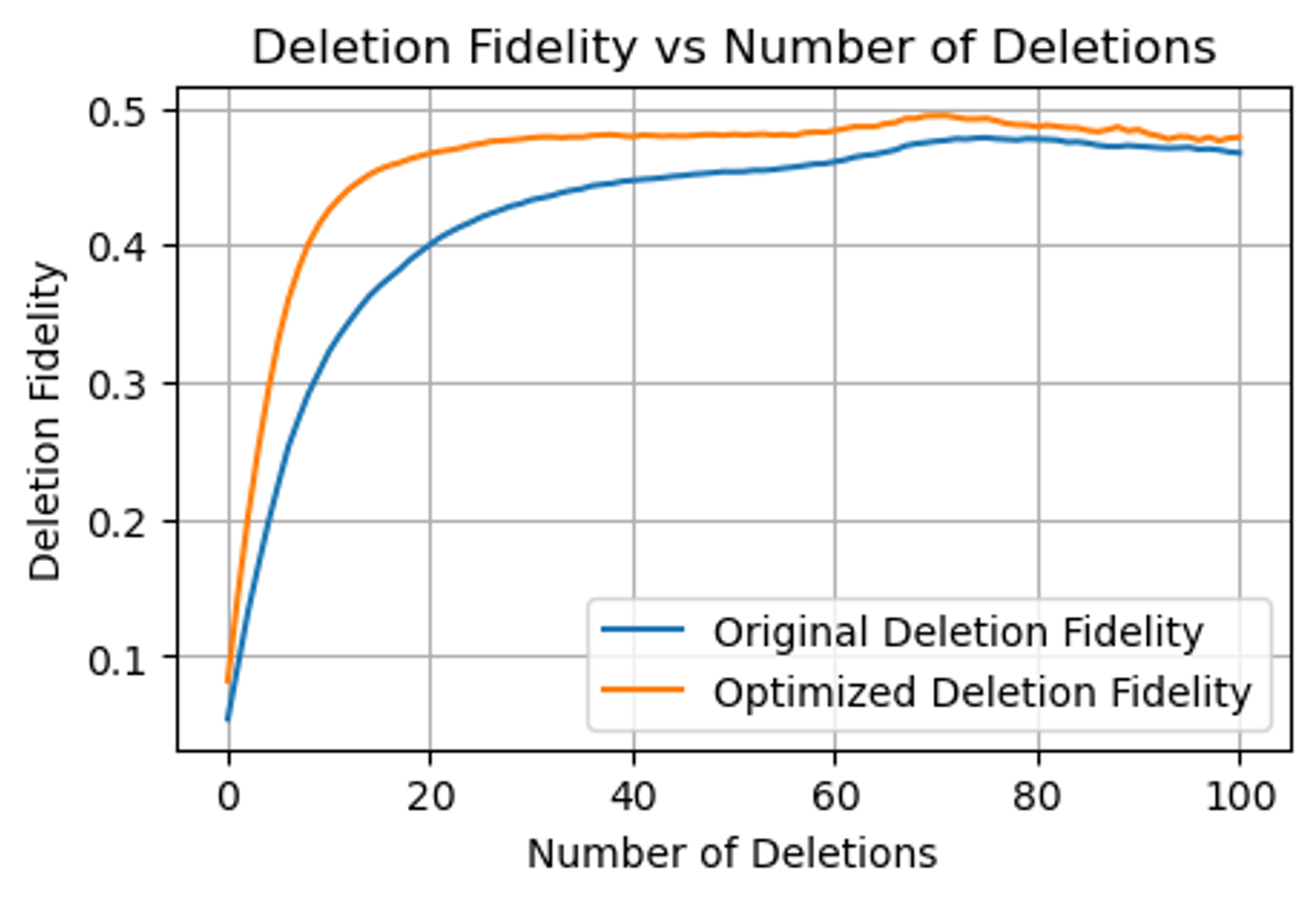}}
    \caption{The deletion fidelity (AOPC\(\uparrow\)) of explanations on the original neighborhood and optimal neighborhood.}
    \label{fig:fid}
\end{figure}

%% file: figures/rq2_2.tex
\begin{figure*}
    \centering
    \resizebox{0.8\textwidth}{!}{\includegraphics{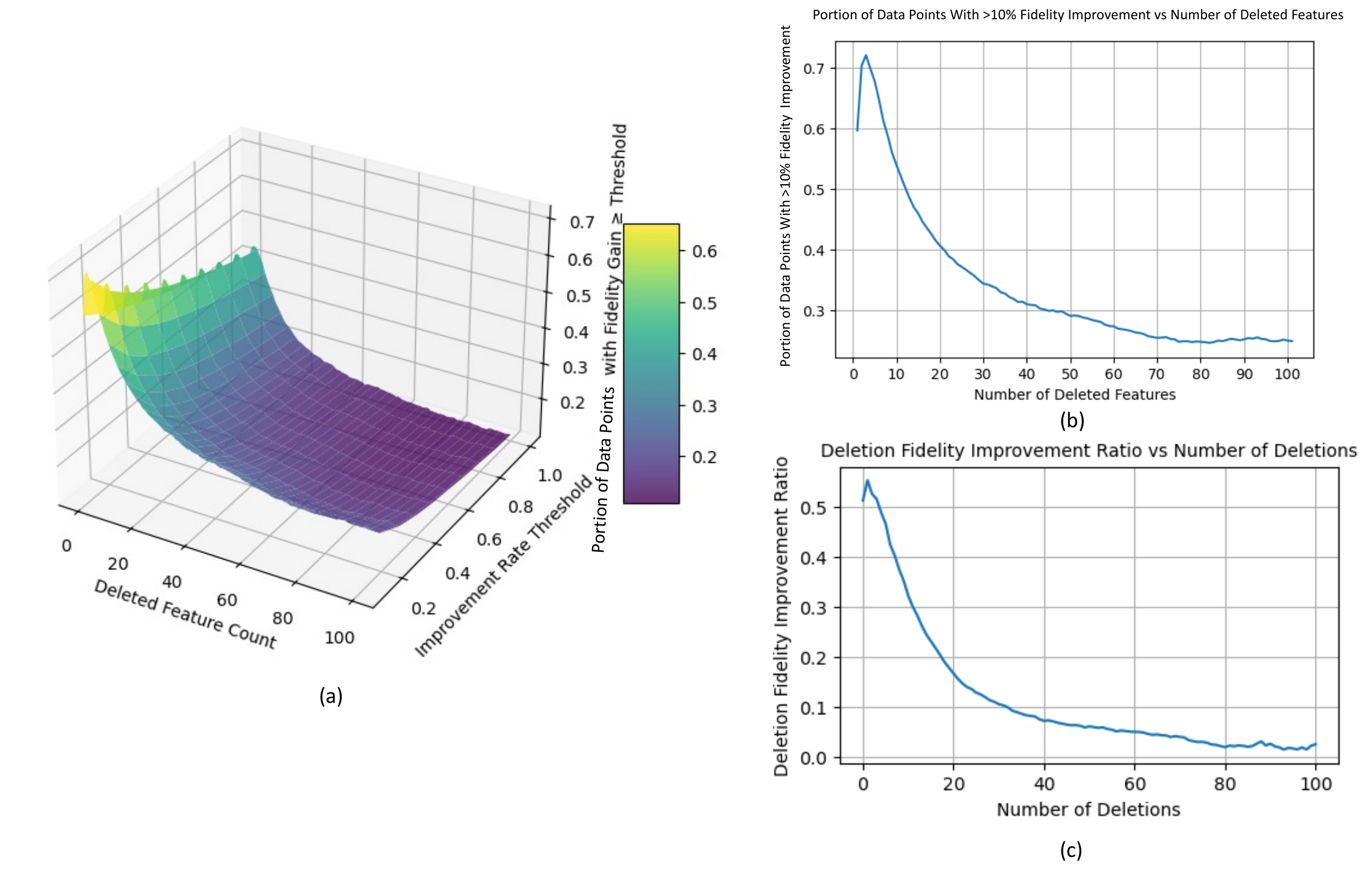}}
    \caption{The fidelity of explanations on the original neighborhood and optimal neighborhood.}
    \label{fig:imp2}
\end{figure*}

%% file: figures/case_study.tex
\begin{figure}[t]
    \centering
    % 调整宽度: width=1.0\linewidth 表示占满通栏宽度
    % 如果图片留白较多，可以改为 width=0.9\linewidth
    \includegraphics[width=\linewidth]{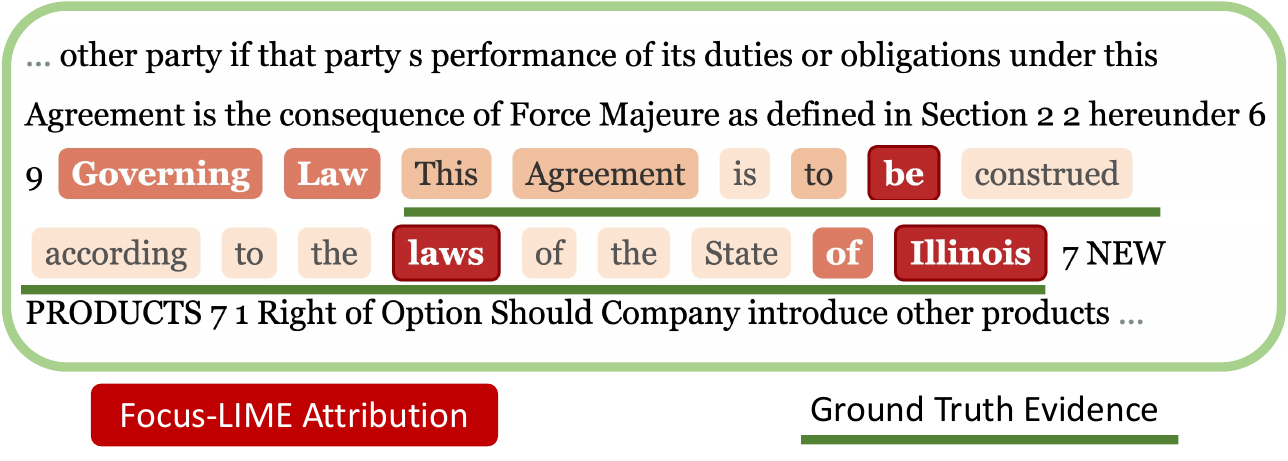}
    
    \caption{
        The importance words identified by Focus-LIME and the ground-truth evidence highlighted by expert lawyers for the "Governing Law" clause checking task.
    }
    \label{fig:case-study}
\end{figure}

%% file: tables/human_alignment.tex
\begin{table}[t]
\centering
\caption{Alignment with human-annotated evidence on \textbf{CUAD}, measured by Recall@k.
Higher values indicate better evidence localization.
GPT-4o is used as the target model and Qwen3-30B-A3B as the proxy model.}
\label{tab:alignment}
\small
\begin{tabular}{lccc}
\toprule
k & 50\% & 150\% & 200\% \\
\midrule
GPT-4o + Qwen3-30B &0.43 & 0.65  & 0.91 \\
GPT-4o + Qwen2.5-14B & 0.40 & 0.62 & 0.87 \\
\bottomrule
\end{tabular}
\end{table}

%% file: chapters/6.conclusion.tex
\section{Conclusion}

In this paper, we introduced \textbf{Focus-LIME}, a framework designed to resolve the critical challenge of surgical word-level attribution dilution in long-context LLMs.
By leveraging a cost-efficient proxy to rigorously curate the search space, our method restores the tractability of surgical interpretation, allowing the target model to focus its sampling budget on the most relevant subspace. 
Extensive experiments demonstrate that Focus-LIME make word-level explanations for long-context LLMs practicable, and has the potential to improve explanation fidelity in more general settings.